\documentclass[conference]{IEEEtran}
\IEEEoverridecommandlockouts
\usepackage{cite}
\usepackage{amsmath,amssymb,amsfonts}
\usepackage{algorithmic}
\usepackage{graphicx}
\usepackage{textcomp}
\usepackage{xcolor}

\usepackage{subfigure}
\usepackage{epsfig}
\usepackage{multirow}
\usepackage{makecell}
\usepackage{bm}
\usepackage{comment}
\usepackage{xcolor}
\usepackage{dsfont}

\usepackage{hyperref}
\hypersetup{hidelinks,
	colorlinks=true,
	allcolors=black,
	pdfstartview=Fit,
	breaklinks=true}

\definecolor{darkgreen}{rgb}{0,0.7,0}
\newcommand{\blue}[1]{{\color{blue}{#1}}}
\newcommand{\red}[1]{{\color{red}{#1}}}
\definecolor{darkyellow}{RGB}{220,180,72}
\newcommand{\yellow}[1]{{\color{darkyellow}{#1}}}
\newcommand{\green}[1]{{\color{darkgreen}{#1}}}
\usepackage{colortbl}
\definecolor{mygray}{gray}{.91}
\newlength\savewidth\newcommand\shline{\noalign{\global\savewidth\arrayrulewidth\global\arrayrulewidth .8pt}\hline\noalign{\global\arrayrulewidth\savewidth}}
\usepackage{pifont}

\def\BibTeX{{\rm B\kern-.05em{\sc i\kern-.025em b}\kern-.08em
    T\kern-.1667em\lower.7ex\hbox{E}\kern-.125emX}}
\begin{document}

\title{Positive Label Is All You Need for Multi-Label Classification\\
\thanks{$^{\ast}$Equal contributions. \quad $^{\dagger}$Corresponding author.}
}

\author{\IEEEauthorblockN{Zhixiang Yuan$^{\ast}$}
\IEEEauthorblockA{
\textit{Anhui University of Technology}\\
Maanshan, China \\
zxyuan@ahut.edu.cn}
\and
\IEEEauthorblockN{Kaixin Zhang$^{\ast}$$^{\dagger}$}
\IEEEauthorblockA{
\textit{Anhui University of Technology}\\
Maanshan, China \\
kxzhang0618@163.com}
\and
\IEEEauthorblockN{Tao Huang}
\IEEEauthorblockA{
\textit{The University of Sydney}\\
Darlington, Australia \\
thua7590@uni.sydney.edu.au}
}

\maketitle

\begin{abstract}
Multi-label classification (MLC) faces challenges from label noise in training data due to annotating diverse semantic labels for each image. Current methods mainly target identifying and correcting label mistakes using trained MLC models, but still struggle with persistent noisy labels during training, resulting in imprecise recognition and reduced performance. Our paper addresses label noise in MLC by introducing a positive and unlabeled multi-label classification (PU-MLC) method. To counteract noisy labels, we directly discard negative labels, focusing on the abundance of negative labels and the origin of most noisy labels. PU-MLC employs positive-unlabeled learning, training the model with only positive labels and unlabeled data. The method incorporates adaptive re-balance factors and temperature coefficients in the loss function to address label distribution imbalance and prevent over-smoothing of probabilities during training. Additionally, we introduce a local-global convolution module to capture both local and global dependencies in the image without requiring backbone retraining. PU-MLC proves effective on MLC and MLC with partial labels (MLC-PL) tasks, demonstrating significant improvements on MS-COCO and PASCAL VOC datasets with fewer annotations. Code is available at: \url{https://github.com/TAKELAMAG/PU-MLC}.
\end{abstract}

\begin{IEEEkeywords}
Multi-label classification, image recognition, positive-unlabeled learning, noisy label
\end{IEEEkeywords}

\section{Introduction}

Recently, multi-label classification (MLC)~\cite{Chen19b, Chen19c, Ridnik21} has gained significant attention as an natural image often contains multiple objects or concepts. Traditional approaches to MLC treat it as a series of binary classification tasks, each determining the presence or absence of individual classes.

\begin{figure}[t]
\begin{center}
\includegraphics[width=0.9\linewidth]{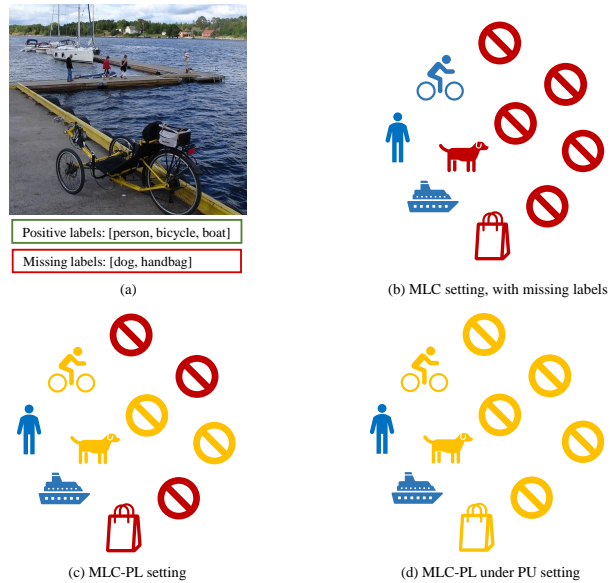}
\end{center}
   \caption{Comparisons of different learning methods in MLC. (a) an image which has two missing labels. To train the sample image, (b) missing labels in traditional MLC methods are mistakenly classified as negative labels; (c) MLC-PL samples a proportion of labels, but still encounters false negative labels; (d) Our method treats all negative labels as unlabeled ones. \blue{Blue}, \red{red}, and \yellow{yellow} icons denote positive, negative, and unknown labels, respectively.
   }
\label{fig:PU_setting}
\end{figure}

Noisy labels, a prevalent issue in MLC datasets due to annotation difficulties~\cite{Ridnik21}, disrupt training and impair performance (see Figure~\ref{fig:PU_setting} (a)-(b)). To address this, certain methods~\cite{Ridnik21} suggest initially training models with these noisy labels, then using the trained model to correct or eliminate mislabeled data. However, the involvement of mislabeled labels in training phase can still negatively influence the process and potentially lead to inaccuracies in identifying noisy labels.

The mislabeling issue becomes more pronounced in the context of multi-label classification with partial labels (MLC-PL) \cite{Huynh20, Chen22, Pu22}. In MLC-PL, where models are trained with partially labeled datasets to minimize annotation costs (refer to Figure \ref{fig:PU_setting} (c)), the limited label information increases the model's vulnerability to label noise. Addressing this, some approaches~\cite{Huynh20} aim to mitigate noisy labels' impact by adjusting the loss weight for each sample. Others explore semantic-aware representations for pseudo label generation~\cite{Chen22} or blend category-specific semantic representations across different images~\cite{Pu22}. However, these MLC-PL methods, like their MLC counterparts, still incorporate mislabeled samples in training. This practice can lead to inaccurate loss weight assessments and pseudo label creation, ultimately impacting model performance.

To address the issue of noisy labels in multi-label classification (MLC) and MLC with partial labels (MLC-PL), which impair training performance, we propose a novel approach in the absence of a reliable method to identify these noisy labels: removing all labels. Drawing inspiration from positive-unlabeled (PU) learning~\cite{Plessis15, Chen20}, which trains classifiers using only positive labels and compares favorably with traditional positive-negative (PN) learning (refer to Figure~\ref{fig:PU_setting}(d)), our method discards all negative labels and relies on positive and unlabeled data for training MLC models. This strategy, leveraging the imbalance of negative labels in MLC datasets (see Figure~\ref{fig:PN_imbalance}), reduces annotation errors. PU learning, known for its robustness and accuracy, especially with noisy negative labels, uses an unbiased risk estimator for better performance. It provides more accurate and informative labeling using soft labels, contrasting with hard labels in conventional methods.

As a result, we introduce a novel method, positive and unlabeled multi-label classification (PU-MLC), adapting PU learning for MLC tasks by integrating multiple binary classifications. To address the significant imbalance between positive and negative labels in MLC, we introduce an adaptive re-balance factor in the PU loss to adjust loss weights effectively. Recognizing the complexity of training multiple binary tasks in MLC compared to standard PU learning, we propose an adaptive temperature coefficient module. This module fine-tunes the sharpness of predicted probabilities in the loss function, preventing over-smoothing in early training stages and enhancing optimization. Additionally, we present a novel local-global convolution module that incorporates both local and global image dependencies. This module enriches existing convolution layers with global information without requiring backbone retraining.

Our PU-MLC method is both simple and effective for MLC and PU-MLC tasks. It demonstrates strong performance even with limited positive labels, reducing annotation costs. Our extensive experiments on benchmark datasets MS-COCO~\cite{Lin14} and PASCAL VOC 2007~\cite{Everingham10} show that PU-MLC significantly improves performance in both MLC and MLC-PL settings, while utilizing fewer annotated labels.

\section{Related Work}

\subsection{Multi-Label Classification}

Multi-label classification (MLC) task aims to recognize semantic categories in a given image, which usually contains multiple objects or concepts. Previous works~\cite{Chen19,Chen19b,You20} propose to construct pairwise statistical correlations using the first-order adjacency matrix obtained by graph convolutional networks (GCN)~\cite{Kipf16}.  Although the above methods achieve noteworthy success, they cannot extract higher-order correlations and can attract overfitting on small training sets. Some works~\cite{Lanchantin21,Zhao21} introduce transformer to extract complicated dependencies among visual features and labels. 

\textbf{MLC with partial labels (MLC-PL).} Traditional multi-label classification (MLC) tasks rely on fully annotated datasets, and making such datasets is expensive, time-consuming, and error-prone. To reduce the cost of annotation, multi-label classification with partial labels (MLC-PL) attempts to train models with partially-annotated labels per image, which both contain positive and negative labels. Recent works~\cite{Durand19,Huynh20,Chen22} propose to generate pseudo labels to those unknown samples based on the learned knowledge in the training model, and then train the model with ground-truth partial labels and generated pseudo labels. 

\subsection{Positive-Unlabeled (PU) learning}

Different from the traditional positive-negative (PN) learning in the binary classification task, PU learning aims to train the model with only positive and unknown labels~\cite{Bekker20}. Recent advances~\cite{Plessis15,Kiryo17,Chen20,huang2022greedynasv2} have achieved remarkable progress in deep learning.  However, these methods rely heavily on the class prior estimation. While the class prior in the training dataset may not always correctly represent the label distribution in the validation set, and thus performing PU learning without class prior becomes an emergent topic~\cite{Chen20,Hu21, Chang21, Gong21}. For example, vPU~\cite{Chen20} proposes a variational principle to achieve superior performance without class prior. In this paper, we extend PU learning to MLC task based on vPU~\cite{Chen20}.

\section{Proposed Approach: PU-MLC}

\begin{figure*}[t]
  \centering
  \subfigure[]{
    \label{fig:framework} 
    \centering 
    \begin{minipage}[c][0.32\linewidth]{0.58\linewidth}
        \centering
        \includegraphics[width=\linewidth]{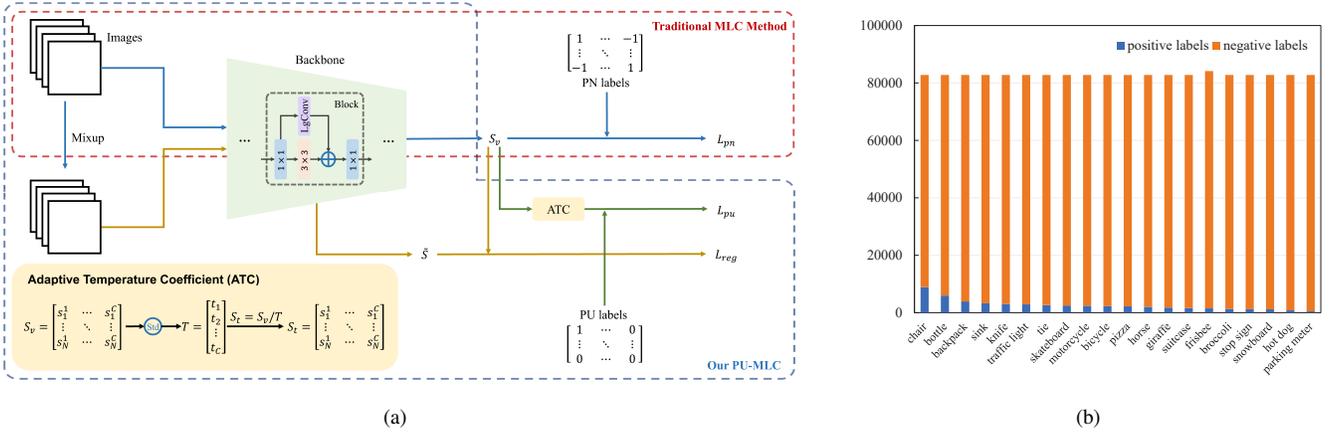}
    \end{minipage}
    
  }
  \hspace{2mm}
  \subfigure[]{
    \label{fig:PN_imbalance}
    \centering
    \begin{minipage}[c][0.32\linewidth]{0.36\linewidth}
    \centering
    \includegraphics[width=0.95\linewidth]{PNImbalance2.eps}
    \end{minipage}
  }
  \caption{(a) Overview of our proposed PU-MLC. Instead of using positive and negative labels in the traditional MLC method (red box), our PU-MLC conducts a positive-unlabeled (PU) learning strategy with only partial positive labels leveraged. Besides, we introduce mixup regularization loss and the adaptive temperature coefficient module to further boost the performance. Additionally, we enhance the global representations in backbone by integrating a local-global convolution module to every $3\times 3$ local convolutions. \textit{Std}: standard deviation. (b) Histograms of the number of positive and negative labels in each category. We randomly select 20 categories from MS-COCO train set.}
  \label{fig:dis}
\label{fig:distributions}
\end{figure*}

\subsection{MLC as PU learning}

\textbf{MLC as PN learning.} MLC task is usually formulated as multiple binary classification sub-tasks, and each sub-task aims to recognize whether a specific category is in the input image.
Formally, for a MLC task with $C$ categories, let $\bm{s}\in\mathbb{R}^{N\times C}$ and $\bm{y}\in\{-1, +1\}^{N\times C}$ be the predicted logits and the ground-truth positive and negative (PN) labels, respectively, where $N$ denotes batch size, the overall classification loss is formulated as
\begin{equation} \label{eq:MLC_Loss1}
\begin{aligned}
\mathcal{L}_{\mathrm{mlc}} =& \frac1{C\times N}
\sum_{c=1}^C \sum_{n=1}^N [\mathds{1}(y_{n,c} = +1) \mathcal{L}_{+}(\sigma(s_{n,c})) \\
&+\mathds{1}(y_{n,c} = -1) \mathcal{L}_{-}(\sigma(s_{n,c}))],
\end{aligned}
\end{equation}
where $\sigma(\cdot)$ is the Sigmoid function, $\mathds{1}(\cdot)$ is an indicator function that takes the value 1 only if the condition is true and 0 otherwise, $\mathcal{L}_+$ and $\mathcal{L}_-$ denote losses on positive and negative labels, respectively. 

Before presenting our PU-learning based MLC method, we first rewrite the learning objective of the above positive-negative (PN) classification loss (\eqref{eq:MLC_Loss1}) as the expected risk on the training set. The total risk $R_\mathrm{mlc}$ is accumulated with all PN sub-tasks, and for each task (category) with the class prior (proportion of positive labels) $\pi_\mathrm{p}$ and $\boldsymbol{S} \in \mathbb{R}^{M}$ being its corresponding logits on the training set with $M$ images, its risk is formulated as
\begin{equation} \label{eq:MLC_risk1}
\begin{aligned}
R_{\mathrm{pn}} &= \pi_\mathrm{p} \mathbb{E}_{\mathcal{P}}[\mathcal{L}_{+}(\sigma(\boldsymbol{S}))]+(1-\pi_\mathrm{p}) \mathbb{E}_{\mathcal{N}}[\mathcal{L}_{-}(\sigma(\boldsymbol{S}))],
\end{aligned}
\end{equation}
where the images regarding to their label types are split into positive set $\mathcal{P}$ and negative set $\mathcal{N}$, and we have the expectations of positive and negative losses
\begin{equation} \label{eq:MLC_risk2}
\begin{aligned}
&\mathbb{E}_{\mathcal{P}}[\mathcal{L}_{+}(\sigma(\boldsymbol{S}))] = \frac{1}{|\mathcal{P}|} \sum_{s_m\in\mathcal{P}} \mathcal{L}_{+}(\sigma(s_m)), \\
&\mathbb{E}_{\mathcal{N}}[\mathcal{L}_{-}(\sigma(\boldsymbol{S}))] = \frac{1}{|\mathcal{N}|}\sum_{s_m\in\mathcal{N}}\mathcal{L}_{-}(\sigma(s_m)).
\end{aligned}
\end{equation}

\textbf{PN to PU.} In this paper, we aim to train a MLC model with only positive labels; \textit{i.e.}, our training set is composed of a positive set $\mathcal{P}$ and an unlabeled set $\mathcal{U}$  (mixture of unlabeled positive and negative images). Nevertheless, the negative labels are unavailable in our PU setting, and therefore we cannot directly optimize \eqref{eq:MLC_risk1} to obtain our model. In order to train a classifier with positive and unknown labels, a classical method uPU~\cite{Plessis15} introduces an unbiased formulation to the PN learning by rewriting the expectation of negative classification loss $\mathbb{E}_{\mathcal{N}}[\mathcal{L}_{-}(\sigma(\boldsymbol{S}))]$ to
\begin{equation} \label{eq:uPU1}
\begin{aligned}
(1-\bm{\pi}_\mathrm{p}) \mathbb{E}_{\mathcal{N}}[\mathcal{L}_{-}(\sigma(\boldsymbol{S}))] &= 
\mathbb{E}_{\mathcal{U}}[\mathcal{L}_{-}(\sigma(\boldsymbol{S}))] \\
&-\bm{\pi}_\mathrm{p} \mathbb{E}_{\mathcal{P}}[\mathcal{L}_{-}(\sigma(\boldsymbol{S}))],
\end{aligned}
\end{equation}
and thus \eqref{eq:MLC_risk1} could be converted to PU format:
\begin{equation} \label{eq:uPU2}
\begin{aligned}
R_{\mathrm{pu}} &= 
\bm{\pi}_\mathrm{p} \mathbb{E}_{\mathcal{P}}[\mathcal{L}_{+}(\sigma(\boldsymbol{S}))] \\
&- \bm{\pi}_\mathrm{p} \mathbb{E}_{\mathcal{P}}[\mathcal{L}_{-}(\sigma(\boldsymbol{S}))] + 
\mathbb{E}_{\mathcal{U}}[\mathcal{L}_{-}(\sigma(\boldsymbol{S}))],
\end{aligned}
\end{equation}

However, the above method easily causes overfitting in deep neural networks and rely heavily on the class prior, and we empirically find that it performs poorly on the multi-label classification task, as the task is more challenging and many categories have very small class priors. Hence, this paper utilizes a recent PU leaning method vPU~\cite{Chen20}, which proposes a novel loss function based on the variational principle to approximate the ideal classifier without the class prior:
\begin{equation} \label{eq:vPU1}
R_\mathrm{var} =
\log{\mathbb{E}_{\mathcal{U}}[\sigma(\boldsymbol{S})]} - 
\mathbb{E}_{\mathcal{P}}[\log{\sigma(\boldsymbol{S})}].
\end{equation}

Hence, for each category $c$, the classification loss becomes
\begin{equation} \label{eq:vPU2}
\begin{aligned}
\mathcal{L}_\mathrm{var}^{(c)} =& 
\log (\frac{1}{|\mathcal{U}^{(c)}_N|} \sum_{s_u \in \mathcal{U}^{(c)}_N} \sigma(s_u)) - \frac{1}{|\mathcal{P}^{(c)}_N|} \sum_{s_p \in \mathcal{P}^{(c)}_N} \log \sigma(s_p),
\end{aligned}
\end{equation}
here $\mathcal{P}^{(c)}_N$ and $\mathcal{U}^{(c)}_N$ denote positive samples and unlabeled samples of category $c$ in each mini-batch, respectively. Note that vPU also introduces a consistency regularization term $\mathcal{L}_\mathrm{reg}^{(c)}$ based on Mixup~\cite{Zhang17}, which alleviates the overfitting problem and increases the robustness in PU learning.

As a result, in our PU-MLC, the traditional MLC loss in \eqref{eq:MLC_Loss1} is replaced with our PU loss, and the overall loss function is formulated as
\begin{equation}
    \mathcal{L}_\mathrm{pu-mlc} = \sum_{c=1}^{C} (\mathcal{L}_\mathrm{var}^{(c)} + \lambda\mathcal{L}_\mathrm{reg}^{(c)}),
\end{equation}
where $\lambda$ is a scalar to balance the losses and we set $\lambda=1$ in all experiments.

Importantly, our approach diverges from traditional PU learning by including all positive samples $\mathcal{P}$ into $\mathcal{U}$. This ensures that $\mathcal{U}$ maintains a label distribution similar to a conventional training set, a critical factor for the effectiveness of PU learning (refer to our ablation studies for further details).

\subsection{Catastrophic Imbalance of Label Distribution}

MLC datasets typically have a far greater number of negative than positive labels, as shown in Figure~\ref{fig:PN_imbalance}. In PU-MLC, where all negative labels are moved to the unlabeled set, there is a significant imbalance in the number of samples affecting the two terms of $\mathcal{L}_\mathrm{var}$ in equation \eqref{eq:vPU2} within each mini-batch. This differs from conventional PU learning where positive and negative samples are equal in batch size. Applying \eqref{eq:vPU2} as is in our method would cause the unlabeled term to overly influence the optimization, leading to suboptimal results in MLC-PL, especially at low known label ratios (e.g., only achieving 51.8\% mAP with 10\% positive labels).

To alleviate the catastrophic imbalance of label distribution, we aim to narrow down the loss weight of unlabeled term to decrease its importance in optimization. Inspired by focal loss~\cite{lin2017focal} and ASL~\cite{Ridnik21}, we propose a re-balance factor to dynamically re-weight the unlabeled loss based on the predicted probabilities on unlabeled samples, and \eqref{eq:vPU2} is reformulated as
\begin{equation} \label{eq:dis1}
\begin{aligned}
\mathcal{L}_\mathrm{var}^{(c)} =& 
p^{\gamma}_c\log (\frac{1}{|\mathcal{U}^{(c)}_N|} \sum_{s_u \in \mathcal{U}^{(c)}_N} \sigma(s_u)) - \frac{1}{|\mathcal{P}^{(c)}_N|} \sum_{s_p \in \mathcal{P}^{(c)}_N} \log \sigma(s_p),
\end{aligned}
\end{equation}
where $p_c^\gamma$ denotes our re-balance factor, with $p_c=\frac{1}{|\mathcal{U}|}\sum_{s_u \in \mathcal{U}}\sigma(s_u)$ being the mean probability of unlabeled samples, and $\gamma$ is used to control the value of the factor. In our experiments, we set larger $\gamma$ for smaller known label ratios, as the imbalance is severer on smaller ratios and we need a smaller weight on unlabeled loss to balance the loss.

\subsection{Adaptive Temperature Coefficient}

In PU learning, the model serves as an estimator for probabilistic evaluations of unlabeled samples, optimizing them via the unlabeled loss term \cite{Bekker20}. However, the task in MLC, which involves learning multiple binary classifiers, is considerably more complex than the single binary classification task in standard PU methods. This complexity results in a slower convergence rate during the early stages of training. Consequently, the predicted probability distribution tends to be over-smooth, reducing the effectiveness of the unlabeled loss.

To adjust the smoothness of probabilistic distribution, we follow \cite{hinton2015distilling} and propose a temperature coefficient $\tau$ to scale the logit values, \textit{i.e.}, $\bm{s}_t = \bm{s} / \tau$, then the $\bm{s}_t$ is fed into the PU loss in place of the original $\bm{s}$.

By setting $\tau < 1$, the probabilistic distribution becomes sharper, providing more meaningful and impactful feedback to the loss function. However, our empirical findings indicate that a fixed temperature coefficient $\tau$ enhances performance only under certain known label ratios and specific datasets (refer to Table 3 in the appendix). For instance, the MS-COCO dataset benefits from $\tau<1$, whereas the PASCAL VOC dataset shows better results with $\tau>1$. This suggests that the optimal $\tau$ varies not only across different datasets but also among different categories within the same dataset, necessitating individual adjustments.

As a result, we propose an adaptive temperature coefficient module to first measure the sharpness of each category in each batch, then apply independent temperatures on each category. Formally, given the predicted logits $\bm{s}$, the sharpness of each category $c$ is measured using the standard deviation of the logits, and then the temperature is obtained by multiplying a scalar $\alpha$ onto the sharpness value, \textit{i.e.,}
\begin{equation}
    \tau^{(c)} = \min(\alpha\cdot\mathrm{Std}(\bm{s}_c), 1).
\end{equation}
We use a minimum function to ensure that the $\tau^{(c)}$ is less than or equal to 1, since we do not want the $\tau^{(c)}$ to exceed 1, which could even exacerbate the over-smooth.

The final PU loss $\mathcal{L}_\mathrm{val}^{(c)}$ becomes
\begin{equation}
\begin{aligned}
\mathcal{L}_\mathrm{var}^{(c)} =& p^{\gamma}_c\log (\frac{1}{|\mathcal{U}^{(c)}_N|} \sum_{s_u \in \mathcal{U}^{(c)}_N} \sigma(s_u / \tau^{(c)}))\\
&- \frac{1}{|\mathcal{P}^{(c)}_N|} \sum_{s_p \in \mathcal{P}^{(c)}_N} \log \sigma(s_p / \tau^{(c)}).
\end{aligned}
\end{equation}

Our adaptive temperature coefficient is suitable for different known label ratios and datasets, which could gain consistent improvements. The overall framework of our model is illustrated in Figure~\ref{fig:framework}.

\begin{table*}[t]
\setlength\tabcolsep{2.7mm}
\caption{The comparisons on MS-COCO and VOC 2007 under different known label ratios. Note that our PU-MLC only uses partial positive labels, while other methods train models with the same number of positive labels and additional negative labels. $^*$ indicates the backbone is pretrained by CLIP~\cite{Radford21}. Results except DualCoOp and our method are reported by SARB~\cite{Pu22}.}
\footnotesize
\begin{center}
\begin{tabular}{c|c|c c c c c c c c c|c c c}
\shline
Datasets & Methods & 10\% & 20\% & 30\% & 40\% & 50\% & 60\% & 70\% & 80\% & 90\% & \thead{Avg.\\mAP} & \thead{Avg.\\OF1} &  \thead{Avg.\\CF1}\\
\hline
\multirow{6}*{MS-COCO}
& ASL~\cite{Ridnik21} & 69.7 & 74.0 & 75.1 & 76.8 & 77.5 & 78.1 & 78.7 & 79.1 & 79.7 & 76.5 & 46.7 & 47.9 \\
& CL~\cite{Huynh20} & 26.7 & 31.8 & 51.5 & 65.4 & 70.0 & 71.9 & 74.0 & 77.4 & 78.0 & 60.7 & 61.9 & 48.3 \\
& Partial BCE~\cite{Huynh20} & 61.6 & 70.5 & 74.1 & 76.3 & 77.2 & 77.7 & 78.2 & 78.4 & 78.5 & 74.7 & 74.0 & 68.8 \\
& SST~\cite{Chen22} & 68.1 & 73.5 & 75.9 & 77.3 & 78.1 & 78.9 & 79.2 & 79.6 & 79.9 & 76.7 & - & - \\
& SARB~\cite{Pu22} & 72.5 & 76.0 & 77.6 & 78.7 & 79.6 & 79.8 & 80.0 & 80.5 & 80.8 & 78.4 & 76.8 & 72.7 \\
\rowcolor{mygray}
& PU-MLC & \textbf{75.7} & \textbf{78.6} & \textbf{80.2} & \textbf{81.3} & \textbf{82.0} & \textbf{82.6} & \textbf{83.0} & \textbf{83.5} & \textbf{83.8} & \textbf{81.2} & \textbf{77.4} & \textbf{75.7}  \\
& DualCoOp$^*$~\cite{Sun22} & 78.7 & 80.9 & 81.7 & 82.0 & 82.5 & 82.7 & 82.8 & 83.0 & 83.1 & 81.9 & 78.1 & 75.3 \\
\rowcolor{mygray}
& PU-MLC$^*$ & \textbf{80.2} & \textbf{83.2} & \textbf{84.4} & \textbf{85.6} & \textbf{85.9} & \textbf{86.6} & \textbf{87.0} & \textbf{87.1} & \textbf{87.5} & \textbf{85.3} & \textbf{81.7} & \textbf{79.1}  \\
\hline
\multirow{6}*{VOC 2007}
& ASL~\cite{Ridnik21} & 82.9 & 88.6 & 90.0 & 91.2 & 91.7 & 92.2 & 92.4 & 92.5 & 92.6 & 90.5 & 41.0 & 40.9 \\
& CL~\cite{Huynh20} & 44.7 & 76.8 & 88.6 & 90.2 & 90.7 & 91.1 & 91.6 & 91.7 & 91.9 & 84.1 & 83.8 & 75.4 \\
& Partial BCE~\cite{Huynh20} & 80.7 & 88.4 & 89.9 & 90.7 & 91.2 & 91.8 & 92.3 & 92.4 & 92.5 & 90.0 & 87.9 & 84.8 \\
& SST~\cite{Chen22} & 81.5 & 89.0 & 90.3 & 91.0 & 91.6 & 92.0 & 92.5 & 92.6 & 92.7 & 90.4 & - & - \\
& SARB~\cite{Pu22} & 85.7 & 89.8 & 91.8 & 92.0 & 92.3 & 92.7 & 92.9 & 93.1 & 93.2 & 91.5 & \textbf{88.3} & 86.0 \\
\rowcolor{mygray}
& PU-MLC & \textbf{88.0} & \textbf{90.7} & \textbf{91.9} & \textbf{92.0} & \textbf{92.4} & \textbf{92.7} & \textbf{93.0} & \textbf{93.4} & \textbf{93.5} & \textbf{92.0} & 88.2 & \textbf{86.5} \\
& DualCoOp$^*$~\cite{Sun22} & 90.3 & 92.2 & 92.8 & 93.3 & 93.6 & 93.9 & 94.0 & 94.1 & 94.2 & 93.2 & 86.3 & 84.2 \\
\rowcolor{mygray}
& PU-MLC$^*$ & \textbf{91.3} & \textbf{92.9} & \textbf{93.3} & \textbf{93.7} & \textbf{93.8} & \textbf{94.3} & \textbf{94.5} & \textbf{94.6} & \textbf{94.8} & \textbf{93.7} & \textbf{89.8} & \textbf{88.2} \\
\shline
\end{tabular}
\end{center}
\label{table:mAP}
\end{table*}

\begin{figure}[t]
\begin{center}
   \includegraphics[width=0.9\linewidth]{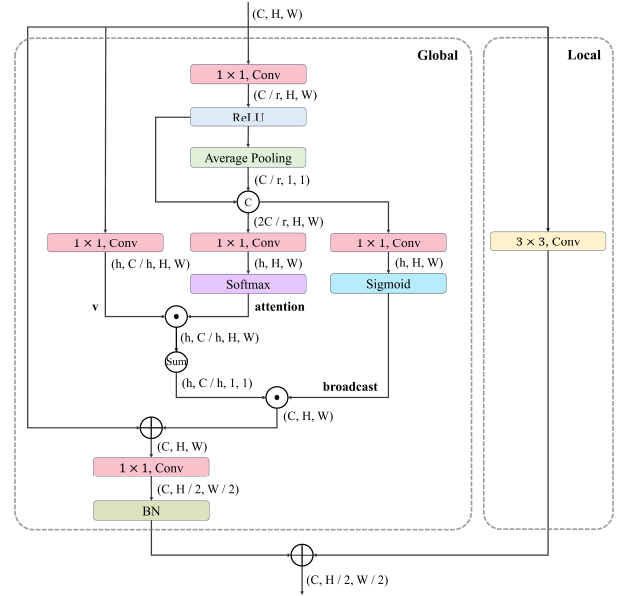}
\end{center}
   \caption{Illustration of Local-global convolution.}
   \label{fig:LgConv}
\end{figure}
\subsection{Local-Global Convolution}

Vision transformers \cite{dosovitskiy2020image, liu2021swin} have shown notable advancements over classical CNNs by capturing global dependencies, yet they face challenges like high memory usage, deployment difficulties, and limitations in lightweight models. Addressing these, we introduce a convolution-based global module, LgConv, designed as a plug-and-play enhancement for CNNs without necessitating retraining of the backbone.

As illustrated in Figure~\ref{fig:LgConv}, LgConv augments traditional local convolution with a global branch. This branch first transforms input features to incorporate both local and global information (via average pooling), followed by two $1\times1$ convolutions creating spatial multi-head attentions and a broadcast attention. Softmax and Sigmoid functions are then applied. The process concludes with a $1\times1$ convolution and batch normalization to project the feature.

To preserve the pretrained backbone's semantic integrity, we initialize the global branch's scale parameters $\gamma$ in the final batch normalization layer at a minimal value (0.0001). This ensures the global branch's initial influence on original features is minimal, allowing for a smooth evolution of the backbone during training.

\nocite{huang2022knowledge}
\nocite{huang2022dyrep}

\section{Experiments}

To verify the efficacy of PU-MLC, we conduct extensive experiments on two popular benchmarks MS-COCO~\cite{Lin14} and PASCAL VOC~\cite{Everingham10}. We adopt similar training strategies following previous works~\cite{Chen22,Pu22}, which will be detailedly discussed in appendix.

\subsection{Results on MS-COCO}

\textbf{MLC-PL setting.} To demonstrate the effectiveness of the PU-MLC, we compare our PU-MLC with current published state-of-the-art methods. As the experimental results shown in Table~\ref{table:mAP}, our PU-MLC significantly outperforms previous methods under different known label ratios. For example, on a high known label ratio of 90\%, we obviously surpass SARB by 3.0\% in mAP.  Compared with previous methods, our method achieves state-of-the-art results in average mAP, OF1 and CF1, which are 81.2\%, 77.4\% and 75.7\%, respectively. DualCoOp uses CLIP~\cite{Radford21}, a large-scale vision-language pre-trained model, as its backbone to achieve exceptional performance. For a fair comparison, by only using the same visual model, our method achieves superior performance than DualCoOp with both visual and language models. 

Note that these significant improvements are obtained with even fewer annotated labels used in training compared to other methods (\textit{e.g.}, with 10\% known label ratio, we only use 10\% positive labels, while other methods use 10\% positive labels and 10\% negative labels), this indicates that our method is more effective and efficient on limited training annotations. As shown in Table~\ref{table:label_numbers}, the number of annotated labels used by PU-MLC in model training is much smaller than other methods based on PN.  Concretely, our method achieves the best results while decreasing the amount of annotated labels by 96.4\% at each known label ratio.  

\begin{table}[t]

\caption{Comparisons of the number of annotated labels used in training on MS-COCO. \textit{Reduction}: the reduction ratio on used training annotations of our method compared to others.}
\setlength\tabcolsep{0.5mm}
\footnotesize
\begin{center}
\begin{tabular}{c|c c c|c c c}
\shline
\multicolumn{1}{c|}{\multirow{2}{*}{Methods}} & \multicolumn{3}{c|}{PU-MLC} & \multicolumn{3}{c}{Others}\\ \cline{2-7}
 & 10\%  & 50\% & 100\% & 10\%  & 50\% & 100\% \\
\hline
Positive & 24,103 & 120,517 & 241,035 & 24,103 & 120,517 & 241,035\\
Negative & 0 & 0 & 0 & 638,160 & 3,190,802 & 6,381,605 \\
Total & 24,103 & 120,517 & 241,035 & 662,263 & 3,311,319 & 6,622,640\\
\hline
Reduction & \green{-96.4\%} & \green{-96.6\%} & \green{-96.4\%} & - & - & -\\
\shline
\end{tabular}
\end{center}
\label{table:label_numbers}
\end{table}

\begin{table}[h]
\caption{mAP on MS-COCO in MLC setting.}
\setlength\tabcolsep{4mm}
\footnotesize
\begin{center}
\begin{tabular}{c|c c c}
\shline
Methods & mAP & OF1 &  CF1 \\
\hline
ResNet-101~\cite{He16} & 77.3 & 76.8 & 72.8 \\
Cop~\cite{Wen20} & 81.1 & 75.1 & 72.7 \\
CADM~\cite{Chen19c} & 82.3 & 79.6 & 77.0 \\
ML-GCN~\cite{Chen19b} & 83.0 & \textbf{80.3} & 78.0 \\
\rowcolor{mygray}
PU-MLC & \textbf{84.2} & 79.1 & \textbf{78.2} \\
\shline
\end{tabular}
\end{center}
\label{table:MLC}
\end{table}
\textbf{MLC setting.} Since our method is designed for both MLC and MLC-PL tasks, we also conduct experiments to validate our performance on traditional MLC. As shown in Table~\ref{table:MLC}, we achieve promising performance compared to previous methods. Similar to MLC-PL, our method in MLC is trained with only positive labels, and discards a large number of negative labels (negative labels are $\sim26.5\times$ more than positive labels), our results can still outperform those methods trained with full annotations. Besides, compared with our PN learning baseline ResNet-101, our MLC-PL significantly outperforms it by 6.9\% in mAP, which demonstrates that our method is beneficial to MLC setting by ignoring those noisy negative labels.

\subsection{Results on Pascal VOC 2007}
Table~\ref{table:mAP} shows the comparisons between PU-MLC and state-of-the-art methods on Pascal VOC. Although Pascal VOC has a small size of the sample and simple categories, and many previous methods achieve splendid results, we still outperform them on average mAP and CF1. Especially on the most challenging 10\% known labels, we obviously surpass SARB by 2.3\% in mAP. On high known label ratios, our improvements are not as significant as that in MS-COCO dataset, a possible reason is that VOC dataset is much easier and smaller than MS-COCO, and using the previous methods can also obtain impressive performance. Additionally, we compare our method with DualCoOp. By using only the same visual model, our approach achieves improvements across all the known label ratios.

\section{Conclusion}
In this paper, we propose positive and unlabeled multi-label classification (PU-MLC). By removing all the negative labels in training, our method benefits from the cleaner annotations. Besides, we introduce an adaptive re-balance factor and adaptive temperature coefficient to better adapt PU learning in MLC task, which achieves significant improvements, especially on small known label proportions. Finally, we design a local-global convolution module to effectively capture both local and global dependencies within the image. Extensive experiments on MS-COCO and PASCAL VOC datasets demonstrate our efficacy. Adopting more advanced PU learning methods and combining recent approaches on model architectures in MLC would be a potential direction of improving PU-MLC.

\bibliographystyle{IEEEtran}
\bibliography{icme2023template}

\end{document}